%% file: main.tex
\documentclass[conference]{IEEEtran}

\usepackage[T1]{fontenc}
\usepackage{amsmath,amssymb,amsfonts}
\usepackage{amsthm}
\usepackage{graphicx}
\usepackage{tikz}
\usepackage[numbers]{natbib}
\usepackage{url}
\usepackage{enumitem}
\usepackage{algorithm}
\usepackage{algorithmic}
\usepackage{tcolorbox}
\usepackage{booktabs}
\usepackage{array}
\usepackage{subcaption}
\usepackage[hidelinks]{hyperref}

\newtheorem{proposition}{Proposition}

\newtcolorbox{highlight}{
  colback=blue!4!white,
  colframe=blue!30!gray!60,
  boxrule=0.4pt,
  arc=1.5pt,
  left=6pt, right=6pt,
  top=4pt, bottom=4pt
}

\begin{document}

\title{When Agents Play Differently: Localizing Human--Agent Behavioral
Divergence with Conditional Distributional Testing}

\author{%
\IEEEauthorblockN{Oleg Smirnov\IEEEauthorrefmark{1},
John Pertoft\IEEEauthorrefmark{1},
Sara Karimi\IEEEauthorrefmark{1}\IEEEauthorrefmark{2}}
\IEEEauthorblockA{%
\IEEEauthorrefmark{1}King AI Labs, Microsoft Gaming, Stockholm, Sweden\\
\IEEEauthorrefmark{2}KTH Royal Institute of Technology, Stockholm, Sweden}
}

\maketitle

\begin{abstract}
Evaluating whether artificial agents behave like humans requires comparing not
just average outcomes but entire behavioral distributions across diverse
contexts. We propose a stratified framework that computes nonparametric
discrepancy statistics within contextual cells, and discovers subgroup rules
identifying where agent behavior departs from human play. We show that
session-level metrics dominated by win/loss outcomes offer little signal beyond
mean comparisons, whereas process-level metrics such as move efficiency reveal
genuine behavioral differences. Applied to a commercial match-3 game, it uncovers patterns
invisible to aggregate metrics: a playtesting bot is diverging substantially
from humans in certain game modes, while showing no detectable divergence in
others. We release our implementation as open source to support reproducibility
and adoption.
\end{abstract}


\input{01_introduction}

\input{02_related_work}

\input{03_formulation}

\input{04_method}

\input{05_setup}

\input{06_results}

\input{07_conclusion}

\bibliographystyle{IEEEtranN}
\bibliography{main}

\end{document}

%% file: 01_introduction.tex
\section{Introduction}
The evaluation of artificial agents hinges on their ability to behave in a
\emph{human-like} manner. In interactive domains such as digital games,
human-likeness is not merely aesthetic: agents that emulate human play enable
scalable testing, balance evaluation, and user modeling without the cost of
large-scale human
trials~\citep{yannakakis2018artificial,holmgaard2018automated}. Conversely,
agents that diverge sharply from human strategies can produce misleading test
results or exploit unrealistic gameplay patterns. A bot validated only on
aggregate outcomes such as overall win rate may still play a level in a
distinctly non-human way, leading a designer to mis-estimate its difficulty or
misjudge how players will experience it.

Existing metrics often rely on aggregate performance measures, e.g., average
scores or success rates, which summarize only the mean
outcome~\citep{swiechowski2025many}. Such scalar indicators obscure
differences in variability, timing, or decision structure that define how
humans actually behave: two agents may exhibit identical mean success rates
yet differ fundamentally in the distribution or dynamics of their actions. A
more rigorous formulation views human-likeness as a problem of
\emph{distributional alignment}: given observations from humans and agents
under shared contexts, assess whether their behavioral distributions
coincide. This connects directly to nonparametric two-sample hypothesis
testing~\citep{ho2016generative,torabi2018behavioral}

\begin{figure}[h]
\centering
\includegraphics[width=\linewidth]{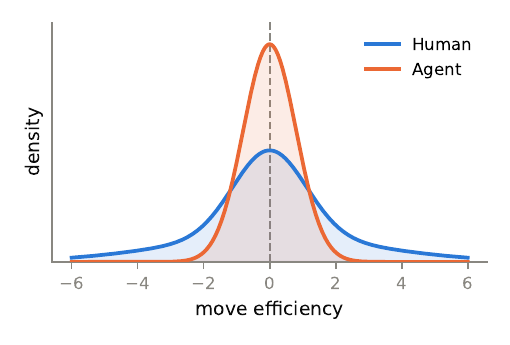}
\caption{Illustrative example. Two groups with the same mean move efficiency but
  different distributions: human play is more variable and heavier-tailed than
  the more consistent agent. A mean comparison reports no difference, whereas a
  distributional test detects it.}
\label{fig:mean-blind}
\end{figure}

In practice, however, such comparisons must be made \emph{conditionally} on
relevant covariates, such as content or player segments, rather than on pooled
data. Imbalance in the frequency of particular contexts can otherwise falsely
appear as behavioral divergence. Moreover, when the null hypothesis of a
global two-sample test is rejected, it is often useful to conduct \emph{a
local two-sample test} to identify specific regions where the two
distributions differ~\citep{kim2019global}. We address these challenges with a
framework for quantifying behavioral alignment between agents and humans
across contextual segments, building on meta-analysis and nonparametric
two-sample testing to construct a diagnostic map of where and how agent
behavior departs from that of humans. Where believability has classically been
assessed as a global verdict, such as whether an observer can tell an agent
from a human, our approach instead localizes where that resemblance breaks
down. Because the resulting map is expressed as human-readable conditions on
context variables, it tells a designer which agent to trust on which content
and turns a diffuse ``not human-like enough'' into specific, localized targets
for improvement.

Our contributions are threefold: (1) a conditional formulation of
behavioral alignment that compares full distributions within matched contexts,
(2) a diagnostic pipeline combining cell-level discrepancy estimation,
stabilization, and interpretable subgroup discovery, and (3) a
production-scale case study on a match-3 game, showing that the inferred
divergence depends strongly on both agent type and behavioral metric.

%% file: 02_related_work.tex
\section{Related Work}
\label{sec:related_work}

\paragraph{Human-likeness and playtesting in games}
Believability and human-likeness in games have long been assessed through
Turing-test-style judgments and believability
competitions~\citep{livingstone2006turing,hingston2012believable,togelius2012assessing}.
Recent work quantifies human-likeness in specific behaviors such as navigation
with learned or statistical
evaluators~\citep{colbert2022human,swiechowski2025many}, while learned agents
are increasingly used for automated playtesting, from procedural personas to
deep-learning policies~\citep{holmgaard2018automated,gudmundsson2018human}. Such
approaches typically render a global verdict, human-like or not, or optimize
agents toward it. We instead localize where and how an agent's behavior
diverges from human play, conditional on game context.

\paragraph{Distributional two-sample testing}
Comparing full distributions rather than means is well studied: Wasserstein
distance \citep{villani2008optimal,peyre2019computational}, energy distance
\citep{szekely2004testing}, and kernel MMD \citep{gretton2012kernel} each
provide nonparametric test procedures with known consistency guarantees.
Classifier-based two-sample tests instead use a trained discriminator's
held-out accuracy as the statistic~\citep{cai2020two,kim2021classification}. We
favor kernel discrepancies, which require no training step and align with
distributional indistinguishability. Our approach treats these as
interchangeable cell-level building blocks rather
than standalone global tests, embedding them within a stratified aggregation
and rule-discovery pipeline.

\paragraph{Stratification and meta-analytic aggregation}
Combining within-stratum evidence into a global test dates to the van~Elteren
and CMH procedures \citep{van1960combination,mantel1959statistical} and to
random-effects meta-analysis with inverse-variance weighting
\citep{dersimonian1986meta}. Our aggregation of cell-level standardized
statistics follows this tradition but extends it in two directions: the
per-cell tests are distributional rather than mean-based, and the aggregated
statistics feed a tree-based rule-discovery step that localizes heterogeneous
effects.

\paragraph{Multiple testing over discovered structure}
Exploring many candidate subgroups demands error control, for which the
Benjamini-Hochberg procedure provides false-discovery-rate (FDR) guarantees
\citep{benjamini1995controlling}. We apply depth-wise correction within tree
levels and control false discoveries over the data-selected partition by
multi-split aggregation of held-out $p$-values \citep{meinshausen2009pvalues}.
The goal of localizing where two distributions differ, rather than only
testing globally, is formalized by \citet{kim2019global}, who propose a
two-stage procedure that first tests a global null and then identifies regions
of significant divergence. Our rule tree serves a similar purpose but
discovers interpretable subgroup structure through recursive partitioning
rather than post-hoc region detection.

%% file: 03_formulation.tex
\section{Problem Formulation and Metric Choice}
\label{sec:formulation}
Let $Y$ denote an outcome of interest and let $X$ be a random variable
describing context (e.g., level difficulty or game mode) with realizations $x
\in \mathcal{X}$. A mean baseline compares the agent and human populations
within each context,
\begin{equation}
    \Delta(X) = \mathbb{E}[Y^{\text{agent}} \mid X]
              - \mathbb{E}[Y^{\text{human}} \mid X],
\end{equation}
and aggregates $\Delta(X)$ over contexts. Uniform averaging treats all
contexts as equally relevant. A weighted expectation under a target context
distribution $q(X)$,
\begin{equation}
    M(q) = \mathbb{E}_{X \sim q}[\Delta(X)],
\end{equation}
instead aligns the estimand with the production mix or a stress-test
distribution. More fundamentally, the comparison remains limited to \emph{mean}
differences: two groups may exhibit identical means within each context yet
differ in variance, tail behavior, or multimodality
(Figure~\ref{fig:mean-blind}). Capturing such
distributional discrepancies motivates the conditional two-sample testing
approach developed in Section~\ref{sec:method}.

We instantiate the framework on a commercial match-3 game, where each session
is one attempt at a level: the player has a limited number of moves to
complete the level's objective, and the session ends in a win or a loss. The
method is agnostic to the outcome variable~$Y$, requiring only that a valid
discrepancy measure and standard error can be computed within each cell.
However, its \emph{informativeness} depends critically on whether the chosen
metric carries distributional structure beyond what a mean comparison
captures.  Many session-level game metrics are strongly coupled with the
binary session outcome. Any metric approximately determined by the outcome
decomposes as
\begin{equation}\label{eq:outcome-dominated}
    Y \approx f(B) + \varepsilon, \qquad \varepsilon \perp\!\!\!\perp B,
\end{equation}
with $B$ the win indicator. When $\mathrm{Var}(f(B)) \gg
\mathrm{Var}(\varepsilon)$ the distribution of $Y$ is governed almost entirely
by the success rate, and the distributional distance reduces, to first order,
to a function of the success-rate difference. In this regime nonparametric
testing presents no additional advantage.

\begin{highlight}
\textbf{Criterion.} The approach adds value only when the residual
$\varepsilon$ itself differs between groups, i.e., when behavioral divergence
exists beyond what the outcome explains.
\end{highlight}

\begin{algorithm*}[t]
\caption{Conditional Two-Sample Testing Pipeline}
\label{alg:cts}
\begin{algorithmic}[1]
\REQUIRE Dataset with outcome $Y$ and context covariates $X$
\STATE \textbf{Discretize} continuous covariates in $X$ into ordinal bins to define a finite grid of context cells.
\STATE \textbf{Form context cells:} group the first group by unique combinations of $X$. Match the second group by exact values or wildcard (missing) context values.
\FOR{each cell $i$}
    \STATE Compute a two-sample discrepancy statistic $t_i$ and its standard error $s_i$ with analytic or resampling-based variance estimation.
\ENDFOR
\STATE \textbf{(Optional) Partial pooling:} apply a random-effects model to obtain stabilized estimates $(\tilde{t}_i, \tilde{s}_i)$.
\STATE \textbf{Build rule tree:} recursively partition cells to maximize meta-analytic $\chi^2$ gain $(\chi^2_L+\chi^2_R)-\chi^2_{\mathrm{parent}}$, with $\chi^2_l = (\sum_{i\in l} \omega_i \tilde{t}_i)^2 / \sum_{i\in l} \omega_i$ and $\omega_i = 1/\tilde{s}_i^2$.
\STATE \textbf{Control false discoveries:} depth-wise correction within the tree, with formal control over the discovered partition by multi-split aggregation of held-out $p$-values.
\STATE \textbf{Output:} interpretable subgroup rules with effect estimates and significance levels.
\end{algorithmic}
\end{algorithm*}

This reduction can be made exact. Within each cell write $P_g(Y \mid x) =
p_{g,i} F_{\mathrm{win}} + (1-p_{g,i}) F_{\mathrm{lose}}$ for group $g$, with
success rates $p_{g,i}$ and outcome-conditional distributions
$F_{\mathrm{win}}, F_{\mathrm{lose}}$ assumed stable across groups.
\begin{proposition}[Outcome-dominance reduction]\label{prop:outcome}
If $\mathsf{Dist}$ is an integral probability metric with symmetric function
class and the component distributions $F_{\mathrm{win}}, F_{\mathrm{lose}}$ are
identical across groups, then for each cell~$i$,
\begin{equation}\label{eq:od-bound}
    \theta_i = \big|p_{1,i} - p_{2,i}\big| \cdot
    \mathsf{Dist}(F_{\mathrm{win}},\, F_{\mathrm{lose}}).
\end{equation}
\end{proposition}
\begin{proof}
Fix a cell and write $\Delta p = p_{1,i} - p_{2,i}$. Then $P_1(Y \mid x) -
P_2(Y \mid x) = \Delta p\,(F_{\mathrm{win}} - F_{\mathrm{lose}})$, and for any
integral probability metric with symmetric function class $\mathcal{F}$,
$\mathsf{Dist}(P_1, P_2) = \sup_{f \in \mathcal{F}} |\Delta p|\,\big|E_{F_{\mathrm{win}}}[f]
- E_{F_{\mathrm{lose}}}[f]\big| = |\Delta p|\cdot \mathsf{Dist}(F_{\mathrm{win}},
F_{\mathrm{lose}})$.
\end{proof}
The per-cell effect is thus the success-rate gap times a constant $D_0 =
\mathsf{Dist}(F_{\mathrm{win}}, F_{\mathrm{lose}})$ that depends on the domain
but not the agent. Only when the outcome-conditional distributions themselves
differ between groups does distributional testing add signal beyond a
proportions test.

Applying this criterion to common game metrics: win rate is a Bernoulli
variable fully characterized by its mean, so testing it is exactly a
proportions test. Total moves used and game score are bimodal mixtures whose
mode masses track the success rate, and are therefore
outcome-dominated. Attempts-per-success, though not outcome-dominated, is
undefined for deterministic single-rollout agents, which produce a fixed
outcome on a given level. Process-level move-efficiency metrics avoid all
three: they are continuous rather than binary, largely orthogonal to the
win/loss outcome, and defined on every session for both groups regardless of
outcome.

We therefore test two process-level move-efficiency metrics, score per move
and cascades per move. These capture \emph{how} an agent plays rather than
\emph{whether} it succeeds: a winning and a losing session can show similar
per-move efficiency if the player performed well but ran out of moves. Two
agents may achieve identical success rates yet differ in play style, e.g.,
with one proceeding through steady progress, while the other relying on
high-variance \emph{cascades}, where a cascade is a chain reaction in which cleared tiles
trigger further clears without an additional move.

%% file: 04_method.tex
\section{Conditional Two-Sample Testing}
\label{sec:method}
To capture differences beyond the mean, we extend the comparison to two-sample
hypothesis testing within contexts. For each context $x$, let $P_1(Y \mid x)$
and $P_2(Y \mid x)$ denote the conditional outcome distributions of two
groups, such as the agent and humans. A discrepancy measure
$\mathsf{Dist}(\cdot,\cdot)$ is applied within each context and aggregated
under a target distribution $q(X)$:
\begin{equation}
    D(q) = \mathbb{E}_{X \sim q}\!\big[
          \mathsf{Dist}\big(P_1(\cdot \mid X), P_2(\cdot \mid X)\big)\big].
\end{equation}
Any discrepancy yielding an effect estimate and a standard error (SE) is
admissible, such as Wasserstein~\citep{villani2008optimal}, energy
distance~\citep{szekely2004testing}, or kernel
MMD~\citep{gretton2012kernel}. Beyond a global summary, the method localizes
where differences arise by aggregating per-cell standardized statistics
meta-analytically, treating each node as a single degree-of-freedom
inverse-variance--pooled statistic that accounts for unequal variances and
sample sizes across contexts~\citep{dersimonian1986meta,gelman2007data}.

\paragraph{Cells and per-cell testing}
Observations from both groups are merged and continuous covariates discretized
into ordered bins, yielding a grid of context cells. Where a covariate is
undefined for one group, those observations act as wildcards that match all
concrete values. Within each cell, a two-sample test contributes an effect
estimate $t_i$ and standard error $s_i$ as local evidence of divergence. A
well-defined SE is essential for the meta-analytic aggregation that follows,
obtained analytically under asymptotic normality or by resampling. When cells
share second-group observations their statistics are correlated, so bootstrap
resampling is performed jointly: the second group is resampled once per
iteration and filtered to each cell, yielding correct marginal standard
errors.

\paragraph{Partial pooling}
When the context grid is large or imbalanced, per-cell estimates are
stabilized by partial pooling under a random-effects
model~\citep{dersimonian1986meta,gelman2007data}: each $t_i$ is shrunk toward
a global mean $\mu$ with weight $w_i = \tau^2/(\tau^2 + s_i^2)$, giving
$\tilde{t}_i = w_i t_i + (1-w_i)\mu$ and $\tilde{s}_i^2 = w_i s_i^2$. This
mitigates noise in sparse cells while retaining systematic heterogeneity.

\paragraph{Rule discovery}
Stabilized cell statistics are partitioned by a greedy tree inspired by
CART~\citep{breiman2017classification}. At each node, candidate splits are
scored by the improvement in meta-analytic $\chi^2$,
\begin{equation}\label{eq:gain}
    \mathrm{Gain} = (\chi^2_L + \chi^2_R) - \chi^2_{\text{parent}},
\end{equation}
where $\chi^2_l = (\sum_{i\in l} \omega_i \tilde{t}_i)^2 / \sum_{i\in l}
\omega_i$ with $\omega_i = 1/\tilde{s}_i^2$ is the squared
inverse-variance--weighted pooled effect within node~$l$. To see why this is the right objective, note that under the alternative the
aggregated statistic follows a noncentral $\chi^2$ with noncentrality
$\lambda = \sum_i \theta_i^2/\sigma_i^2$, the signal available to the test.
Splitting a cell into children $a$ and $b$ raises $\lambda$ by
$\tfrac{n_a n_b}{n_a+n_b}(\theta_a-\theta_b)^2/V$, which coincides with the gain
in~\eqref{eq:gain} when child effects are homogeneous. The tree therefore
greedily maximizes the detectable signal, splitting where effects differ most
and adding a degree of freedom only when the gain outweighs it. The resulting rule tree yields
human-readable subgroup definitions.

\paragraph{False-discovery control}
Because many hypotheses are tested across the tree, we apply depth-wise
Benjamini-Hochberg or Benjamini-Yekutieli
correction~\citep{benjamini1995controlling,benjamini2001control} at each depth
level, pruning non-significant branches. Because the tree is selected from
data, these depth-wise node $p$-values are descriptive. Formal false-discovery
control over the discovered partition is provided by a multi-split procedure,
in which held-out leaf $p$-values are aggregated across repeated player-level
splits by the adaptive quantile rule of \citet{meinshausen2009pvalues} with a
Benjamini-Yekutieli correction across cells.

Algorithm~\ref{alg:cts} summarizes the complete procedure. The method's validity rests on an
adequate discretization of continuous covariates and on ignorability, the
assumption that all variables affecting both group membership and the outcome
are among the conditioning covariates.

%% file: 05_setup.tex
\section{Experimental Setup}
\label{sec:setup}
We apply the conditional two-sample testing pipeline to session-level data
from a commercial match-3 game, comparing two playtesting bots (automated
agents that play levels in place of human testers) against human players: a
supervised behavioral cloning model (SL) trained on the production data
inspired by the deep-learning playtesting approach of
\citet{gudmundsson2018human}, and a heuristic model (Hbot) implementing
handcrafted hint-based strategies. Both agents are evaluated against the same
human population across 24 context cells defined by level difficulty (0--3)
and game mode: Chain (cascade-oriented), Clear (objective clearing), and Route
(stream routing, with three variants). The three modes differ in objective:
Chain rewards cascades, Clear requires removing designated target tiles within
the move limit, and Route requires guiding board elements to fixed
destinations. For each agent, the dataset comprises approximately $110{,}000$
player sessions and $5{,}000$ agent sessions obtained via a table sample of
the full dataset. Two continuous move efficiency metrics are tested: score per
move and cascades per move. Both metrics exhibit low outcome dominance
($R^2(\text{win}) = 0.036$ for score per move and $0.121$ for cascades per
move), confirming that their distributional content is not reducible to
success rate differences.

The pipeline utilizes euMMD~\citep{bodenham2023eummd} as
a computationally efficient discrepancy measure. Per-cell $p$-values use
default 200 permutations. The two-sided $p$-value is floored at $0.5/201
\approx 0.0025$, which is the $p \approx 0.002$ saturation seen in most
cells. Per-cell standard errors are computed separately from 100 distinct
bootstrap resamples. Partial pooling is applied via DerSimonian-Laird random
effects, and rule discovery uses $\chi^2$ gain with depth-wise
Benjamini-Hochberg correction at $\alpha=0.05$. The main analysis focuses on
these two agents, which exhibit qualitatively different divergence profiles.

%% file: 06_results.tex
\section{Experimental Results}
\label{sec:results}

\subsection{Per-Cell Two-Sample Tests}
\begin{figure}[t]
\centering
\includegraphics[width=\linewidth]{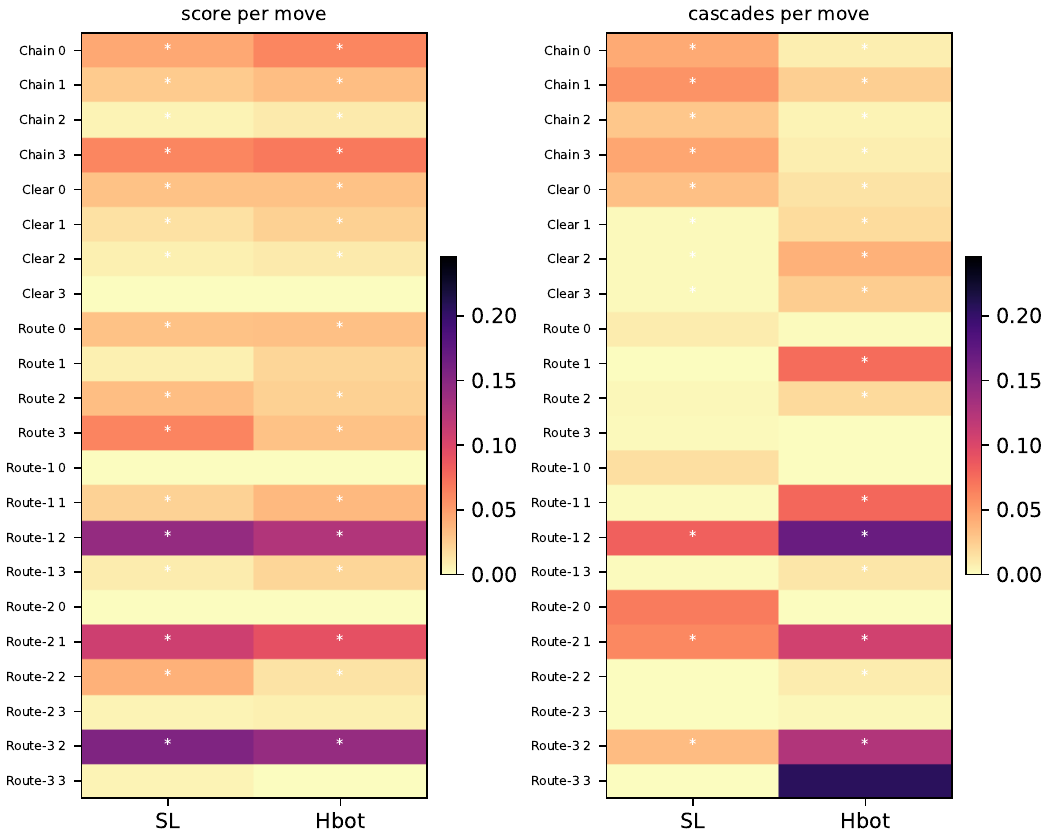}
\caption{Per-cell MMD effect estimates ($\Delta$) for SL and Hbot on score per
  move and cascades per move, across the 24 context cells (difficulty $\times$
  game mode). Color intensity encodes effect magnitude; ${}^*$ marks cells
  significant at $\alpha=0.05$ after BH correction. The cascades panels show the
  agent-dependent dissociation: SL diverges most in Chain, Hbot most outside
  it.}
\label{fig:heatmap}
\end{figure}

Figure~\ref{fig:heatmap} reports per-cell effect estimates for both agents on
both metrics. Here $\Delta$ is the unbiased MMD$^2$ $U$-statistic and can be
negative under the null, and $p$-values are obtained by permutation. For score
per move, the two agents show similar gradients: large effects in Chain at all
difficulties and at medium-difficulty Route variants, with hard Clear emerging
as the most aligned subgroup for both ($\Delta \approx 0.000$--$0.001$,
non-significant). For cascades per move, the per-cell pattern differs
structurally between the two agents. The SL model diverges consistently in
Chain across all difficulties ($\Delta \approx 0.03$--$0.05$), with only
isolated larger effects elsewhere. The heuristic model shows the opposite
emphasis, with Chain effects roughly half the SL magnitudes ($\Delta \approx
0.006$--$0.025$) and substantially larger divergence in Clear and Route
variants at higher difficulties ($\Delta \approx 0.02$--$0.17$). Several
small-sample cells exhibit large standard errors relative to their point
estimates. Partial pooling shrinks these noisy estimates without distorting
well-estimated cells.

\begin{figure*}[t]
\centering
\includegraphics[width=\textwidth]{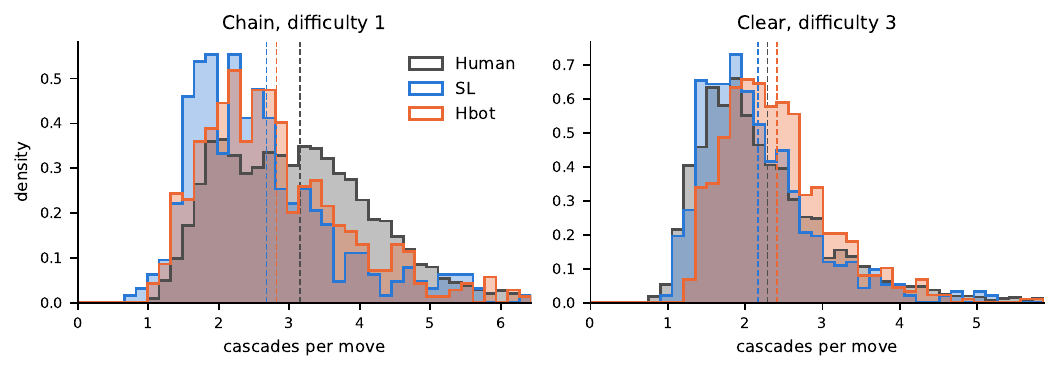}
\caption{Distributions of cascades per move for humans and the two agents in a
  strongly divergent cell (left) and a near-aligned cell (right). Dashed lines mark group means. In Chain the agents concentrate
  at lower cascade rates and rarely sustain the long chains humans produce, and
  in hard Clear the three distributions nearly coincide.}
\label{fig:dist-overlay}
\end{figure*}

\subsection{Discovered Rules}
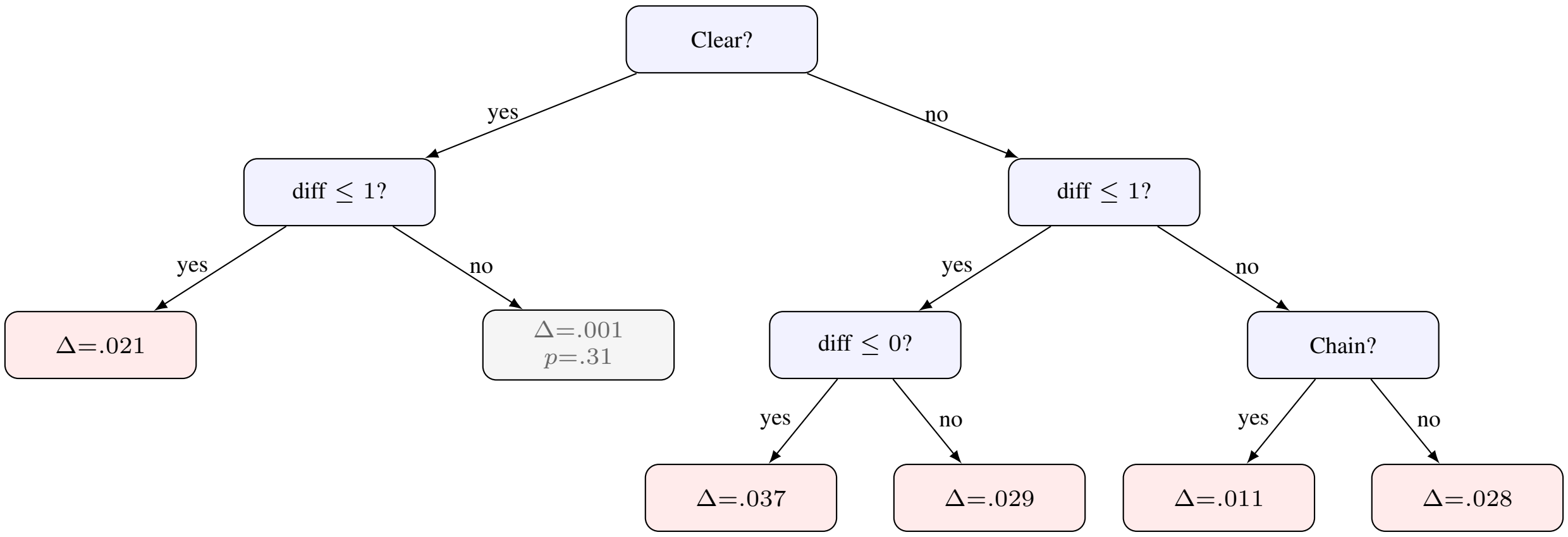
\begin{figure*}[t]
\centering
\begin{tikzpicture}[
    every node/.style={align=center, font=\scriptsize},
    level 1/.style={sibling distance=80mm, level distance=16mm},
    level 2/.style={sibling distance=50mm, level distance=16mm},
    level 3/.style={sibling distance=26mm, level distance=16mm},
    edge from parent/.style={draw, -latex},
    sig/.style={draw, rounded corners, fill=red!8, minimum width=20mm, minimum height=7mm},
    nsig/.style={draw, rounded corners, fill=gray!8, minimum width=20mm, minimum height=7mm, text=black!60},
    split/.style={draw, rounded corners, fill=blue!5, minimum width=20mm, minimum height=7mm},
]
\node[split] {Clear?}
    child { node[split] {diff $\leq 1$?}
        child { node[sig] {$\Delta{=}.021$}
            edge from parent node[left, font=\scriptsize] {yes} }
        child { node[nsig] {$\Delta{=}.001$\\$p{=}.31$}
            edge from parent node[right, font=\scriptsize] {no} }
        edge from parent node[left, font=\scriptsize] {yes}
    }
    child { node[split] {diff $\leq 1$?}
        child { node[split] {diff $\leq 0$?}
            child { node[sig] {$\Delta{=}.037$}
                edge from parent node[left, font=\scriptsize] {yes} }
            child { node[sig] {$\Delta{=}.029$}
                edge from parent node[right, font=\scriptsize] {no} }
            edge from parent node[left, font=\scriptsize] {yes}
        }
        child { node[split] {Chain?}
            child { node[sig] {$\Delta{=}.011$}
                edge from parent node[left, font=\scriptsize] {yes} }
            child { node[sig] {$\Delta{=}.028$}
                edge from parent node[right, font=\scriptsize] {no} }
            edge from parent node[right, font=\scriptsize] {no}
        }
        edge from parent node[right, font=\scriptsize] {no}
    };
\end{tikzpicture}
\caption{Discovered rule tree for SL on score per move. Red nodes are
  significant at $\alpha=0.05$ after depth-wise BH correction; gray is
  non-significant; $\Delta$ is the pooled effect. The tree splits first on
  Clear, then on difficulty, isolating hard Clear as the aligned subgroup. The
  Hbot tree is structurally identical except for the deepest non-Clear split, a
  Route variant in place of Chain.}
\label{fig:tree-score}
\end{figure*}

After partial pooling ($\hat\tau^2 = 3.27 \times 10^{-4}$ and $3.54 \times
10^{-4}$ for SL and Hbot on score per move; $\hat\tau^2 = 1.12 \times 10^{-4}$
and $2.45 \times 10^{-4}$ on cascades per move), the rule discovery procedure
summarizes the cell-level evidence as interpretable subgroup trees. For score
per move the trees for the two agents are structurally similar
(Figure~\ref{fig:tree-score}): both split
first on Clear versus the remaining modes, then on difficulty, and both
isolate hard Clear as a non-significant subgroup ($\Delta = 0.001$, $p = 0.31$
for SL; $\Delta = 0.002$, $p = 0.19$ for Hbot). The trees differ only in the deepest non-Clear split,
Chain for SL and a Route variant for Hbot.

\begin{figure*}[t]
\centering
\begin{minipage}[t]{0.48\textwidth}
\centering
\subcaption{SL}\label{fig:tree-cascades-sl}
\begin{tikzpicture}[
    every node/.style={align=center, font=\scriptsize},
    level 1/.style={sibling distance=42mm, level distance=14mm},
    level 2/.style={sibling distance=20mm, level distance=14mm},
    level 3/.style={sibling distance=25mm, level distance=14mm},
    edge from parent/.style={draw, -latex},
    sig/.style={draw, rounded corners, fill=red!8, minimum width=18mm, minimum height=7mm},
    nsig/.style={draw, rounded corners, fill=gray!8, minimum width=18mm, minimum height=7mm, text=black!60},
    split/.style={draw, rounded corners, fill=blue!5, minimum width=18mm, minimum height=7mm},
]
\node[split] {Chain?}
    child { node[split] {diff $\leq 1$?}
        child { node[sig] {$\Delta{=}.034$\\$p{<}10^{-11}$}
            edge from parent node[left, font=\scriptsize] {yes}
        }
        child { node[sig] {$\Delta{=}.027$\\$p{<}10^{-9}$}
            edge from parent node[right, font=\scriptsize] {no}
        }
        edge from parent node[left, font=\scriptsize] {yes}
    }
    child { node[split] {diff $\leq 0$?}
        child { node[sig] {$\Delta{=}.019$\\$p{<}10^{-5}$}
            edge from parent node[left, font=\scriptsize] {yes}
        }
        child { node[sig] {$\Delta{=}.003$\\$p{=}6{\times}10^{-4}$}
            edge from parent node[right, font=\scriptsize] {no}
        }
        edge from parent node[right, font=\scriptsize] {no}
    };
\end{tikzpicture}
\end{minipage}%
\hfill
\begin{minipage}[t]{0.48\textwidth}
\centering
\subcaption{Hbot}\label{fig:tree-cascades-hbot}
\begin{tikzpicture}[
    every node/.style={align=center, font=\scriptsize},
    level 1/.style={sibling distance=42mm, level distance=14mm},
    level 2/.style={sibling distance=20mm, level distance=14mm},
    level 3/.style={sibling distance=20mm, level distance=14mm},
    edge from parent/.style={draw, -latex},
    sig/.style={draw, rounded corners, fill=red!8, minimum width=18mm, minimum height=7mm},
    nsig/.style={draw, rounded corners, fill=gray!8, minimum width=18mm, minimum height=7mm, text=black!60},
    split/.style={draw, rounded corners, fill=blue!5, minimum width=18mm, minimum height=7mm},
]
\node[split] {Chain?}
    child { node[split] {diff $\leq 1$?}
        child { node[sig] {$\Delta{=}.013$\\$p{<}10^{-4}$}
            edge from parent node[left, font=\scriptsize] {yes}
        }
        child { node[sig] {$\Delta{=}.007$\\$p{=}.008$}
            edge from parent node[right, font=\scriptsize] {no}
        }
        edge from parent node[left, font=\scriptsize] {yes}
    }
    child { node[split] {Route?}
        child { node[nsig] {$\Delta{=}.008$\\$p{>}.05$}
            edge from parent node[left, font=\scriptsize] {yes}
        }
        child { node[split] {Route-2?}
            child { node[sig] {$\Delta{=}.014$\\$p{<}10^{-3}$}
                edge from parent node[left, font=\scriptsize] {yes}
            }
            child { node[sig] {$\Delta{=}.023$\\$p{<}10^{-23}$}
                edge from parent node[right, font=\scriptsize] {no}
            }
            edge from parent node[right, font=\scriptsize] {no}
        }
        edge from parent node[right, font=\scriptsize] {no}
    };
\end{tikzpicture}
\end{minipage}
\caption{Discovered rule trees for cascades per move. Red nodes are
  significant at $\alpha=0.05$ after depth-wise BH correction; gray nodes are
  non-significant; $\Delta$ is the pooled effect. Both agents split first on
  Chain but assign opposite emphasis: SL (a) routes its largest effects to
  Chain, while Hbot (b) routes them to objective-driven content outside it.}
\label{fig:tree-cascades}
\end{figure*}
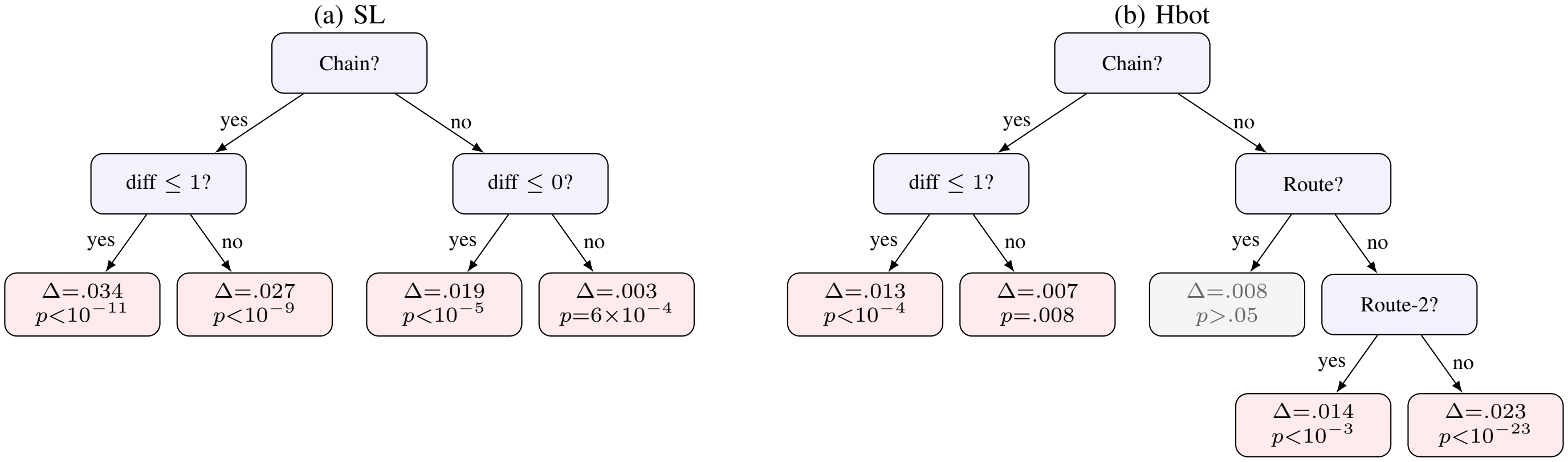

For cascades per move, the trees diverge in emphasis. The SL tree
(Figure~\ref{fig:tree-cascades-sl}) splits first on Chain versus all other
modes, isolating Chain as the dominant divergence axis ($\Delta = 0.034$ at
difficulty $\leq 1$, $\Delta = 0.027$ at difficulty $> 1$). Outside Chain only
a small effect remains, mainly at difficulty 0. The Hbot tree
(Figure~\ref{fig:tree-cascades-hbot}) also splits first on Chain but assigns
substantially smaller effects there ($\Delta = 0.013$ and $\Delta = 0.007$)
and routes its largest effect to objective-driven content ($\Delta = 0.023$).

\subsection{What the Divergence Looks Like}
The rule trees identify \emph{where} the agents diverge, whereas the underlying
distributions show \emph{how}. Figure~\ref{fig:dist-overlay} overlays the human
and agent distributions of cascades per move in the most divergent Chain cell
and in a near-aligned Clear cell. In Chain, human play spreads toward high
cascade rates, with substantial mass beyond three cascades per move, whereas
both agents concentrate at lower values and rarely sustain the long chains that
humans produce, with the learned SL model the most shifted. The gap is one of
both location and shape, and it is precisely the behavior a cascade-oriented
mode is meant to elicit. In hard Clear the three distributions nearly coincide,
the distributional counterpart of the non-significant effect estimate there.

\subsection{Analysis}
The main finding is a contextual dissociation between the two agents on
cascades per move. Although both models cover the full context, their cascade
divergence concentrates on opposite sides of the Chain axis: the SL model is
most non-human in Chain levels dominated by cascade reactions, while
the heuristic model is most non-human outside Chain, such as objective-driven levels
where hints are tailored for targeted completion. A single global mean over
cells averages these effects to a small positive number for both agents,
hiding the structural contrast. The pipeline recovers the contrast as the
dominant feature of the cascade rule trees and makes it explicit at the cell
level. Notably, this discovered contrast is between learning from human demonstrations versus
handcrafted rules, rather than between architectures or objectives within the
learning-based class. While both agents were originally designed to be
human-like, SL by learning from human play and Hbot through hints tuned to
guide play as a human would, each of them departs from human behavior in a
different, context-localized way that an aggregate score would report as
near-alignment.

A secondary finding is a model-independent alignment subgroup: hard Clear
shows no detectable divergence for either agent on score per move ($\Delta
\approx 0.001$, non-significant), and this pattern replicates across both
score-tree variants. Establishing practical equivalence rather than mere
non-significance is possible through a calibrated human self-comparison.

To confirm that the pipeline localizes genuine structure rather than noise, we
validate it on a synthetic benchmark with a divergent region planted
outcome-orthogonally, so that group win rates and overall metric means are
matched and only the within-outcome shape differs. Table~\ref{tab:validation} reports the
outcome: under a global null the full pipeline is well calibrated (realized
FDR $0.004$), whereas the same tree driven by a mean statistic inflates to
$0.216$. On the planted region the MMD pipeline recovers the divergent cells
(precision $0.94$, recall $1.00$) while the mean-only and success-rate
baselines miss them almost entirely. The discovery thus reflects
distributional structure invisible to outcome-level comparison.

\begin{table}[t]
\caption{Synthetic validation: realized false-discovery rate under a global
  null, and recovery of an outcome-orthogonally planted divergent region.
  MMD-based methods recover the signal that mean and success-rate baselines
  are blind to. Nominal $\alpha=0.05$.}
\label{tab:validation}
\centering
\small
\begin{tabular}{l|r|rr}
\toprule
Method & FDR$_{H_0}$ & Precision & Recall \\
\midrule
Full pipeline (MMD${+}$tree) & $.004$ & $.94$ & $1.00$ \\
Per-cell MMD ${+}$ BH        & $.000$ & $.97$ & $0.98$ \\
Mean-only (same tree)        & $.216$ & $.31$ & $0.13$ \\
Success-rate ${+}$ BH        & $.046$ & $.41$ & $0.00$ \\
\bottomrule
\end{tabular}
\end{table}

\subsection{Implications}
The two metrics exercise complementary capabilities of the diagnosis. On score
per move the framework certifies where the agents are human-like: both are
largely interchangeable, with hard Clear a shared alignment subgroup. On
cascades per move it instead localizes where they are not, and does so
differently for each agent. This is directly actionable: the maps identify
which agent to trust on which content, and the concentrated cascade gap is a
specific, localized target for improvement rather than a diffuse average
deficit. Because the rules are read off the tree as human-readable conditions
on difficulty and mode, each finding translates into a statement a designer can
act on.

%% file: 07_conclusion.tex
\section{Conclusion and Future Work}
\label{sec:conclusion}
This work proposes the evaluation of human-likeness based on conditional distributional alignment across contexts rather than aggregate outcome
matching. By framing
human-likeness as a stratified two-sample testing problem, the framework provides both a global discrepancy measure and an interpretable rule
structure describing where agents diverge from human play. Applied to a
large-scale commercial game, the analysis reveals that an agent diverges from
human play in game modes dominated by specific mechanics while showing no
detectable divergence in others, a structured pattern that an aggregated score would obscure.

Beyond diagnosis, the per-cell discrepancy statistics and rule tree define a
structured error signal that can further inform targeted finetuning, curriculum
design, or reward shaping. However, such optimization requires care: directing an agent
to close a localized gap can invoke Goodhart's law, raising the targeted metric
without making the underlying behavior more
human-like~\citep{manheim2019categorizing}, so the signal is best used to inform
the training objective rather than to define it.

Among the method limitations, finer context partitions or alternative
metrics could reveal differentiation that the present analysis does
not surface. Moreover, the formal false-discovery control over the discovered partition currently rests on the
multi-split procedure.

The concentration of cascade divergence in specific contexts raises the
question of whether the agent lacks spatial understanding needed to
reproduce human cascade behavior. We leave to future work a broader study
across agent architectures and training regimes and controlled interventions, such as varying the
model's capacity, that would test those hypotheses directly.